\documentclass{article}

\usepackage[preprint]{abha_style_clean}

\usepackage[utf8]{inputenc}
\usepackage[T1]{fontenc}
\usepackage{hyperref}
\usepackage{url}
\usepackage{booktabs}
\usepackage{amsfonts}
\usepackage{nicefrac} 
\usepackage{microtype}
\usepackage{xcolor}
\usepackage{amsmath}
\usepackage{mathrsfs}
\usepackage{graphicx}
\usepackage{multirow}
\usepackage{wrapfig}
\title{Attention-Bayesian Hybrid Approach to Modular Multiple Particle Tracking}

\author{%
  Piyush Mishra\textsuperscript{1,2}, Philippe Roudot\textsuperscript{2}\\
  \textsuperscript{1}Aix Marseille Univ, CNRS, I2M, Turing Centre for Living Systems\\
  \textsuperscript{2}Aix Marseille Univ, CNRS, Centrale Med, Institut Fresnel, Turing Centre for Living Systems\\
  \texttt{piyush.mishra@univ-amu.fr},  \texttt{philippe.roudot@univ-amu.fr} \\
}

\begin{document}

\maketitle

\begin{abstract}
    Tracking multiple particles in noisy and cluttered scenes remains challenging due to a combinatorial explosion of trajectory hypotheses, which scales super-exponentially with the number of particles and frames. The transformer architecture has shown a significant improvement in robustness against this high combinatorial load. However, its performance still falls short of the conventional Bayesian filtering approaches in scenarios presenting a reduced set of trajectory hypothesis. This suggests that while transformers excel at narrowing down possible associations, they may not be able to reach the optimality of the Bayesian approach in locally sparse scenario. Hence, we introduce a hybrid tracking framework that combines the ability of self-attention to learn the underlying representation of particle behavior with the reliability and interpretability of Bayesian filtering. We perform trajectory-to-detection association by solving a label prediction problem, using a transformer encoder to infer soft associations between detections across frames. This prunes the hypothesis set, enabling efficient multiple-particle tracking in Bayesian filtering framework. Our approach demonstrates improved tracking accuracy and robustness against spurious detections, offering a solution for high clutter multiple particle tracking scenarios.
\end{abstract}

\section{Introduction}

\begin{wrapfigure}{r}{0.4\textwidth}
  \centering
  \vspace{-2em}
  \includegraphics[width=\linewidth]{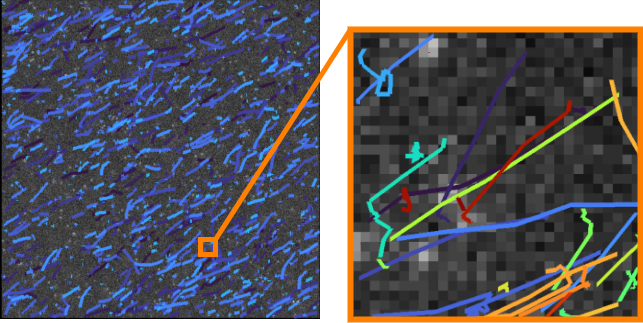}
  \caption{Example highlighting a patch.}
  \label{fig:inset_intro}
  \vspace{-2em}
\end{wrapfigure}

Fluorescence imaging in live cells has uncovered environments in which particles move through crowded and noisy intracellular spaces (\citet{saxton1997single}, \citet{manley2008high}). Within these large fields of view, the cellular behavior can be governed by highly active processes carried out by a dense population of molecules in a limited sub-compartment of the cell (\citet{sage2019super}, \citet{chen2014single}, \citet{david2019augmin}). Although spatially limited, these patches (Fig. \ref{fig:inset_intro}) capture a microenvironment in which interactions and movements of interest are concentrated. Tracking particles in such patches has remained a fundamental challenge that has hindered the understanding of the physiological and pathological functions of cells. (\citet{roudot2013noise}, \citet{chenouard2014objective}, \citet{persson2013extracting}).

A key part of tracking particles at this scale is associating individual detections (\citet{basset2015adaptive}, \citet{smal2009quantitative}, \citet{mantes2024spotiflow}) across frames to reconstruct particle trajectories. Suppose that we have a film with \(T \in \mathbb{N}\) frames. \(\mathbf{Z}_t\) denotes the set of detections in a two-dimensional space \(\zeta = (x, y) \in \mathbb{R}^{1 \times 2}\), at time \(t \in [1, T]\). Each element of \(\mathbf{Z}_t\) can be assembled in a tabular form such that for \(n\) detections throughout the span of the entire film,
\begin{equation}
    \mathbf{Z} = \bigcup_{t=1}^T\{(\zeta, t), \forall \zeta \in \mathbf{Z}_t \} \in \mathbb{R}^{n \times 3}
\end{equation}
The motive, from this data structure, is to have the output \(\hat{\mathbf{X}}^i\) for the \(i^{th}\) trajectory, containing all the predicted positions \(\chi = (\hat{x}, \hat{y}) \in \mathbb{R}^{1 \times 2}\) and their corresponding times of arrival \(t\), such that, for \(B\) predicted number of trajectories,
\begin{equation}
    \hat{\mathbf{X}} = \bigcup_{i=1}^{B}\{(\chi, t, i), \forall (\chi, t) \in \hat{\mathbf{X}}^i\} \in \mathbb{R}^{n \times 4}
\end{equation}

 Conventionally, Bayesian filtering (\citet{reid1979algorithm}, \citet{maybeck1982stochastic}) has provided an iterative framework to solving this problem, maintaining a posterior over the unobserved true states (positions and velocities) \(\mathbf{X}_t\) given the observations \(\mathbf{Z}_{1:t}\).
\begin{equation}
    p(\mathbf{X}_t|\mathbf{Z}_{1:t}) \propto p(\mathbf{Z}_t|\mathbf{X}_t) \int p(\mathbf{X}_t|\mathbf{X}_{t-1}) p(\mathbf{X_{t-1}|\mathbf{Z}_{1:t-1}})d\mathbf{X}_t
\end{equation}
 However, these detections are often noisy and spurious (\citet{smal2008particle}, \citet{chenouard2014objective}), making reliable data association a central difficulty - estimating \(p(\mathbf{X}_t|\mathbf{Z}_{1:t})\) requires considering all possible associations between the measurements in \(\mathbf{Z}\) and the true trajectories in \(\mathbf{X}\), which results in exponentially increasing hypotheses (\citet{fortmann1983sonar}, \citet{reid1979algorithm}). This requires us to effectively prune the hypothesis set to select the best possible set of trajectories.
 
In the past, this problem has been approached by combining a modeling of intracellular dynamics (\citet{jaqaman2008robust}, \citet{roudot2017piecewise}) to estimate the likelihood of trajectory-to-measurement associations and discrete optimization to greedily select the best associations over time (\citet{jaqaman2008robust}, \citet{chenouard2013multiple}, \citet{liang2014novel}, \citet{smal2015quantitative}). Further, such optimization strategies have also been coupled with deep learning strategies like RNNs (\citet{spilger2020recurrent}, \citet{spilger2021deep}) and transformers (\citet{zhang2023motion}) for motion modeling. While RNNs are particularly well-suited in modeling the dynamics of particle trajectories, the sequential nature of their architecture prevents them from being used as an effective strategy for hypothesis handling. Transformers (\citet{vaswani2017attention}), by virtue of their ability to handle broader contexts, could be exploited to learn the associations between input elements (encoder), and to auto-regressively predict motion (decoder). \citet{mishra2024comparative} have provided a proof of concept for this argument using an encoder-decoder vanilla transformer on a regime of 2-particle system following a drifting standard Brownian motion. On comparing it with an elementary Bayesian filtering approach, multiple hypothesis tracking (MHT) (\citet{reid1979algorithm}), they find that attention is more robust to increasing number of hypothetical trajectories, generated through the increase of measurement noise. This suggests that as the number of hypotheses increases, attention learns the underlying patterns in the data to associate each point in the point-cloud to a correct trajectory label. But they also find that MHT is optimal when the number of hypotheses is small or the hypothesis-set is already pruned (with knowledge of a priori). In other recent work, \citet{gallusser2024trackastra} have developed Trackastra which segments trajectories into temporal windows and uses a transformer-based scoring model to evaluate and assemble these into globally consistent tracks. It shows strong performance in general purpose tracking, especially when trained on large and diverse datasets. However, their complete reliance on a transformer structure limits interpretability and adaptability. In wider applications, studies simulating submarine motion modeling like \citet{pinto2022can} and \citet{pinto2023transformer} demonstrate that estimating combined trajectory states can be an effective tracking strategy. With that motivation, \citet{pinto2023transformer}, of note, introduce a modular approach for submarine tracking breaking the task into an association and a filtering problem. The insights from \citet{mishra2024comparative} and the modularity of \citet{pinto2023transformer} provide the foundational motivation of this study.

Thus, it is seen that transformers are robust with large hypothesis sets but can be suboptimal with fewer ones; and are often hard to interpret. Considering these propositions, we argue that combining an attention-based approach for pruning hypotheses and a Bayesian approach for tracking using those pruned hypotheses offers a promising solution to accurate tracking of cluttered regimes. As such, we introduce an Attention-Bayesian Hybrid Approach to modular multiple particle tracking (ABHA). We compare it with MHT, which uses the same Kalman filtering model for state estimation, but relies on an independent frame-by-frame association strategy. This comparison will act as a controlled evaluation, highlighting the effect of ABHA's learned association strategy against classical combinatorial methods, coupled with the same motion modeling strategy. This will further help isolate the impact of the core novelty of ABHA. In Section \ref{sec:methods}, we elaborate on the architecture of ABHA, MHT, the performance metrics and the implementation details that we put in place for the comparison. In Section \ref{sec:results}, we present and analyze the results of this comparison. In Section \ref{sec:conclusion}, we conclude our study and establish grounds for future studies.

\section{Methods}
\label{sec:methods}
\subsection{Proposed Methodology for ABHA}
\subsubsection{A point-cloud association technique using self-attention as a way to prune trajectory hypotheses}
\begin{figure}
    \centering
    \includegraphics[width=\linewidth]{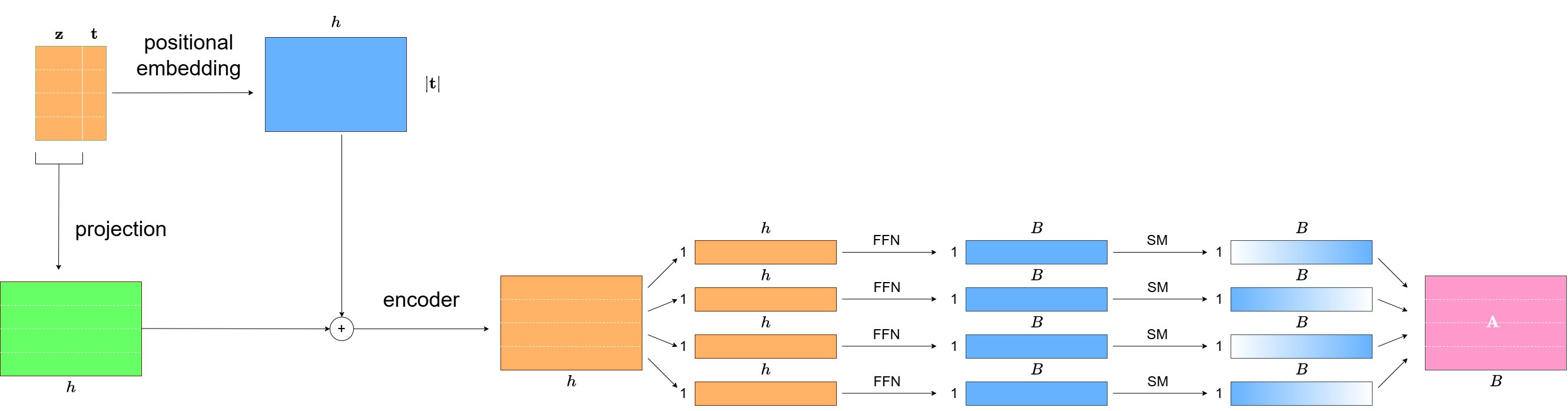}
    \caption{Schematic representation of the association logic. Two data structures \(\mathbf{z}\) and \(\mathbf{t}\) are processed by the transformer encoder to learn associations. Each row of the encoder output represents the latent space of the corresponding measurement, and undergoes feed-forward and softmax layers to output probabilities of association of the corresponding measurement with a label ID. These rows are concatenated to get the association matrix \(\mathbf{A}\).}
    \label{fig:assn_logic}
\end{figure}

The tracking problem is a state estimation problem combined with a combinatorial association problem. If we build a representation for each input and then predict a label for that representation, we can isolate each label for a subsequent Bayesian filtering. This requires considering each data point as input. \(\mathbf{Z}\) can be distributed into two sets which act as the two inputs: measurements, \(\mathbf{z} \in \mathbb{R}^{n\times 2}\) and time, \(\mathbf{t} \in \mathbb{R}^n\) which have corresponding indices. Measurements are projected in a domain of higher dimension \(h\) for \(\mathbf{W} \in \mathbb{R}^{2\times h}\) such that,
\begin{equation}
\begin{split}
    \mathscr{P}:\mathbb{R}^{n\times 2} \rightarrow \mathbb{R}^{n\times h} \\
    \mathbf{z}' = \mathscr{P}(\mathbf{z}) = \mathbf{z}\mathbf{W}
\end{split}
\end{equation}
\(\mathscr{T}\) maps \(\mathbf{t}\) to a set of all of its unique elements,
\begin{equation}
    \mathscr{T}: \mathbb{R}^n \rightarrow \mathbb{S} \text{ where } \mathbb{S} = \{s | s \in \mathbf{t}\}
\end{equation}
Now, this set is projected in the same higher dimension \(h\),
\begin{equation}
\mathscr{L}:\mathbb{S}\rightarrow\mathbb{R}^{|\mathbb{S}|\times h}
\end{equation}
The new representations of measurements and time are combined together such that,
\begin{equation}
\begin{split}
\oplus:\mathbb{R}^{n\times h}, \mathbb{R}^{|\mathbb{S}|\times h} \rightarrow \mathbb{R}^{n\times h} \\
\mathbf{e}^0 = \oplus(\mathbf{z}', \mathbf{t}) = \mathbf{z}' + \mathscr{L}(\mathscr{T}(\mathbf{t}))
\end{split}
\end{equation}
Now this data structure is the input to the encoder, as discussed in Appendix \ref{sec:prerequisites}. Consider \(n_l \in \mathbb{N}\) to be the number of encoder layers such that,
\begin{equation}
\begin{split}
    \mathscr{E}_{n_l}: \mathbb{R}^{n\times h} \rightarrow \mathbb{R}^{n\times h} \\
    \mathbf{e} = \mathscr{E}_{n_l}(\mathbf{e^0})
\end{split}
\end{equation}
Each row of \(\mathbf{e}\) is a latent representation of the corresponding detection. In order to translate it to a set of probabilities for each trajectory label, we pass it through a feed-forward network and a final softmax layer. Hence, each row of \(\mathbf{e}\) goes through a 3-layer feedforward network for \(\mathbf{W}_{f,1}\in\mathbb{R}^{1\times h}, \mathbf{W}_{f,2} \in \mathbb{R}^{h\times h} \text{ and }\mathbf{W}_{f,3}\in\mathbb{R}^{h\times B}\) such that

\begin{equation}
\begin{split}
\mathscr{F}:\mathbb{R}^{1\times h}\rightarrow \mathbb{R}^{1\times B} \\
\mathscr{F}(\mathbf{x}) = \mathbf{x}.\mathbf{W}_{f,1}.\mathbf{W}_{f,2}.\mathbf{W}_{f,3}
\end{split}
\end{equation}
And finally, we get the association matrix \(\mathbf{A} \in \mathbb{R}^{n\times B}\),
\begin{equation}
    \mathbf{A} = \text{concat} \left( \text{softmax} \left( \mathscr{F}(\mathbf{e}_p) \right), \forall p \in [1,n]\right)
\end{equation}
The entire logic is schematized in Fig. \ref{fig:assn_logic}.

\subsubsection{Association Loss Calculation}
We assume that we have a ground truth association matrix \(\mathbf{A}^* \in \mathbb{R}^{n\times \mathbb{T}}\) (where \(\mathbb{T}\) is the total number of trajectories in the ground truth) each of whose columns is representative of a unique trajectory label and each row is a one-hot vector. However, it is likely that the columns of predicted \(\mathbf{A}\) and known \(\mathbf{A}^*\) are not matched, more so if \(B \neq \mathbb{T}\). We want to match the ground truth trajectory to that prediction trajectory which has the highest number of measurements that match it according to \(\mathbf{A}\).

In order to do that, we pad the smaller of the ground truth or prediction matrices with zeros such that \(\mathbf{A}, \mathbf{A}^* \in \mathbb{R}^{n\times \text{max}(B, \mathbb{T})}\). We set a cost matrix \(\mathbf{C} \in \mathbb{R}^{\text{max}(B, \mathbb{T}) \times \text{max}(B, \mathbb{T})}\) such that,
\begin{equation}
\mathbf{C}_{p,q} = - \sum_{r=1}^{\text{max}(B, \mathbb{T})}\mathbf{A}_{r,p}\mathbf{A}^*_{r,q}
\end{equation}
This gives a dissimilarity score between the \(p^{th}\) predicted trajectory and the \(q^{th}\) ground truth trajectory. A more negative dissimilarity score means that the two indices are less dissimilar. Using the Hungarian algorithm (\citet{kuhn1955hungarian}) on \(\mathbf{C}\), we get the reordered column order and apply it to \(\mathbf{A}^*\). Now, we calculate the cross-entropy loss between \(\mathbf{A}\) and the reordered \(\mathbf{A}^{*,o}\),
\begin{equation}
    \mathcal{L}(\mathbf{A}, \mathbf{A}^{*,o}) = -\frac{1}{n}\sum_{p=1}^n \sum_{q=1}^{\text{max}(B, \mathbb{T})}\text{log}(\mathbf{A}_{p,q})\mathbf{A}^{*,o}_{p,q}
\end{equation}
We thus, match the columns of the ground truth association matrix with those of the predicted association matrix to subsequently calculate cross-entropy between them as a metric of loss.

\subsubsection{Assigning associations to their measurements}
\(\mathbf{A}\) is a predicted matrix each of whose rows corresponds to a detection index and each element of a row contains its probability of belonging to a trajectory label denoted by its column index. Now, a predicted class is computed for each data point as:
\begin{equation}
    \mathbf{i} = \text{arg}\max_q\mathbf{A}_{p,q} \forall p \in [1,n]
\end{equation}

Hence \(\mathbf{i} = \begin{bmatrix}i_1 & i_2 & \text{...} & i_n\end{bmatrix}^T \in \mathbb{R}^n\) is the resulting column which is concatenated to \(\mathbf{Z}\) resulting in \(\mathbf{Z}_a\) such that:
\begin{equation}
    \mathbf{Z}_a = \bigcup_{i \in \mathbf{i}}\bigcup_{t \in \mathbf{t}}\bigcup_{\zeta \in \mathbf{z}}\{(\zeta, t, i) \} \in \mathbb{R}^{n \times 4}
\end{equation}
Now, for each trajectory label \(i \in \mathbf{i}\), we extract \(\mathbf{Z}_i \subset \mathbf{Z}_a\).

\subsubsection{Kalman filtering of a pruned tree of trajectory hypotheses}
\label{sec:kalman}
In this section, we adhere to the well-established stochastic filtering technique introduced by \citet{kalman1960new}. Suppose that we are following a particle \(i\) that moves with a constant velocity and its state at time \(t\) is described as \(\mathbf{x}_t^i = \begin{bmatrix}x_t^i & \dot{x}_t^i & y_t^i & \dot{y}_i^t\end{bmatrix}^T\) where \(x_t^i\) and \(y_t^i\) are its positions and \(\dot{x}_t^i\) and \(\dot{y}_t^i\) are its velocity components. Its next state is extrapolated using Kalman filtering with a state transition matrix \(\mathbf{F} = \begin{bmatrix}1 & dt & 0 & 0 \\0 & 1 & 0 & 0 \\0 & 0 & 1 & dt \\ 0 & 0 & 0 & 1\end{bmatrix}\) and the noise covariance matrix of the process \(\mathbf{Q} = \sigma_q^2\begin{bmatrix}\frac{dt^3}{3} & \frac{dt^2}{2} & 0 & 0\\ \frac{dt^2}{2} & dt & 0 & 0\\ 0 & 0 & \frac{dt^3}{3} & \frac{dt^2}{2}\\ 0 & 0 & \frac{dt^2}{2} & dt\end{bmatrix}\) for some process noise parameter \(\sigma_q\) for a sampling time interval \(dt\). See Appendix \ref{sec:process_matrix} for the derivation of \(\mathbf{Q}\). Subsequently, the next state and its uncertainty is updated using the observation matrix \(\mathbf{H} = \begin{bmatrix}1 & 0 & 0 & 0\\ 0 & 0 & 1 & 0\end{bmatrix}\) and the measurement noise covariance matrix \(\mathbf{R} = \sigma_r^2\mathbf{I}_2\) for some measurement noise parameter \(\sigma_r\).

\subsection{Multiple Hypothesis Tracking}
Here, we briefly discuss MHT as the strategy of comparison against ABHA. It uses the same Kalman filtering model as ABHA but uses a purely Bayesian combinatorial association strategy at each frame. At each time step \(t\), a set of active hypotheses \(\mathcal{H}_{t-1}\) is maintained, each representing a predicted object state. Each hypothesis is propagated forward using the Kalman filter as discussed in Section \ref{sec:kalman}, yielding a predicted measurement. Now, to assign observations to hypotheses, a cost matrix \(\mathbf{G}\) is defined each of whose elements contains the Euclidean distance between each predicted and observed measurement. The association is now formulated as a linear assignment problem solved using the Hungarian algorithm (\citet{kuhn1955hungarian}). All assignments beyond a gating threshold \(\tau\) are discarded. Each matched hypothesis is updated using the same Kalman filter as described above. Unmatched hypotheses age incrementally on increasing time steps and are removed if they exceed a maximum allowed number of misses \(a\). Unassigned measurements are used to initialize new hypotheses.

\subsection{Evaluation}
\subsubsection{Data Description and Generation}

We use \(30 \times 30 \) patches from the Viruses dataset from the ISBI Particle Tracking Challenge (\citet{chenouard2014objective}) to compare our approach. This dataset presents the dynamics of diffraction limited spots undergoing stop-and-go motion that mimics viral behavior in vivo. Considering \(N\) particles a ground-truth data structure \(\mathbf{X}\) contains the real positions, \(\mathbf{x}\), of all particles at each time point \(t\). To get a measurement \(\zeta_t^i\), we add a measurement noise to the corresponding state \(\mathbf{x}_t^i\), such that,
\begin{equation}
    \zeta_t^i = \mathbf{x}_t^i + \omega_t^i
\end{equation}
where \(\omega_t^i \sim \mathcal{N}(\mathbf{0}, \sigma_m^2\mathbf{I}_2)\) is the measurement noise.  All these corrupted states \(\zeta_t^i \in \mathbb{R}^{1 \times 2}\) for all time steps \(t\) and all particles \(i\) are stored in a data structure \(\mathbf{Z}_m\) such as:
\begin{equation}
    \mathbf{Z}_m = \bigcup_{i=1}^{N}\bigcup_{t=1}^T\{(\zeta_t^i, t, i) \} \in \mathbb{R}^{N.T \times 4}
\end{equation}

Each tuple \((\zeta, t, i)\) in \(\mathbf{Z}_m\) is independently removed with a false positive probability \(p_{fn}\). This results in \(\mathbf{Z}_{fn}\). The number of false positives at each time step \(t\) follows a Poisson distribution with rate \(\lambda_{fp}\) per unit area. The false positive positions are sampled uniformly over the window of view and are assigned the trajectory -1. This data structure, \(\mathbf{Z}\), is the final data structure, representative of measurements. For data structures used for training, the column \(i\) is kept to construct \(\mathbf{A}^*\), and for validation data, this column is removed.

\subsubsection{Performance Metrics}
\label{sec:perf_metrics}
\textbf{Point-congruence metrics}: A cost function \(d(\mathbf{x}, \chi)\) is defined as the Euclidean distance between ground truth points, \(\mathbf{x} = (x, y)\) and predicted points, \(\chi = (\hat{x}, \hat{y})\). A cost matrix \(\mathbf{D}\) is defined where \(\mathbf{D}_{p,q} = d(\mathbf{x}_p, \chi_q)\). The following minimization problem is solved using the Hungarian algorithm (\citet{kuhn1955hungarian}),
\begin{equation}
    \mathcal{M} = \{(p,q) | \text{arg}\min_{\mathcal{M}}\sum_{p,q}\mathbf{D}_{p,q}, \mathbf{D}_{p,q} \leq d_{\phi} \}
\end{equation}
for some predefined maximum allowable distance, \(d_{\phi}\), for matching. Thus we get \(\mathcal{M}\) as the set of matches between the ground truth and predicted points. Point precision \(P_P\), recall \(R_P\) and Jaccard similarity coefficient, \(JSC_P\) are defined as follows:
\begin{equation}
    P_P = \frac{|\mathcal{M}|}{|\hat{\mathbf{X}}|}
\end{equation}
\begin{equation}
    R_P = \frac{|\mathcal{M}|}{|\mathbf{X}|}
\end{equation}
\begin{equation}
    JSC_P = \frac{|\mathcal{M}|}{|\mathbf{X}| + |\hat{\mathbf{X}}| - |\mathcal{M}|}
\end{equation}

\textbf{Link-congruence metrics}: Sets of links in ground truth and predictions data structures, \(\mathbf{L_X}\) and \(\mathbf{L_{\hat{\mathbf{X}}}}\) are defined as:
\begin{equation*}
\begin{cases}
\mathbf{L_X} = \{ (\mathbf{x}_t^i, \mathbf{x}_{t+1}^i) | (\mathbf{x}_t^i, t, i) \in \mathbf{X},  (\mathbf{x}_{t+1}^i, t+1, i) \in \mathbf{X}\} \\
\mathbf{L_{\hat{X}}} = \{ (\chi_t^j, \chi_{t+1}^j) | (\chi_t^j, t, j) \in \hat{\mathbf{X}},  (\chi_{t+1}^j, t+1, j) \in \hat{\mathbf{X}}\}
\end{cases}
\end{equation*}
Link precision \(P_L\), recall \(R_L\), F1-score \(F1_L\) and Jaccard similarity coefficient \(JSC_L\) are defined as follows:
\begin{equation}
    P_L = \frac{|\mathbf{L_X} \cap \mathbf{L_{\hat{X}}}|}{|\mathbf{L_{\hat{X}}}|}
\end{equation}
\begin{equation}
    R_L = \frac{|\mathbf{L_X} \cap \mathbf{L_{\hat{X}}}|}{|\mathbf{L_X}|}
\end{equation}
\begin{equation}
    F1_L = \frac{2P_LR_L}{P_L + R_L}
\end{equation}
\begin{equation}
    JSC_L = \frac{|\mathbf{L_X} \cap \mathbf{L_{\hat{X}}}|}{|\mathbf{L_X} \cup \mathbf{L_{\hat{X}}}|}
\end{equation}

\textbf{TGOSPA}: We use Trajectory Generalized Optimal Sub-Pattern Assignment, TGOSPA, (\citet{garcia2020metric}) to evaluate the accuracy of predicted trajectories over time. TGOSPA extends the GOSPA (\citet{rahmathullah2017generalized}) metric by aggregating spatial and cardinality errors across all frames. For each frame \(t\), GOSPA (Appendix \ref{sec:gospa}) is computed between the sets of ground truth and predicted positions, and TGOSPA is defined as
\begin{equation}
    TGOSPA = \left( \frac{1}{T} \sum_{t=1}^T d_p^c(\mathbf{X}_t, \hat{\mathbf{X}}_t)^p \right)^{1/p}
\end{equation}
where \(d_p^c\) is the GOSPA distance at time \(t\), \(p\) is the order of the metric, \(c\) is the cutoff threshold. A lower TGOSPA indicates better spatial and temporal consistency in the predicted trajectories.

The dictionary of all parameters for the performance metrics can be found in Appendix Table \ref{tab:perf_params}.

\subsubsection{Implementation Details}
\label{sec:implementation}
\textbf{Method parameters and training details:} We obtain the patches of \((x_{lim}, y_{lim})\) pixels, as defined in Table \ref{tab:method_params} by slicing the dataset with \(SNR = 4\) for low and medium densities from \citet{chenouard2014objective}. The other parameters are also as mentioned in Table \ref{tab:method_params}. The association strategy is trained with 8 such patches and validated on 2 patches. Training is done with a starting learning rate of \(10^{-3}\) and the cyclical annealing strategy of the learning rate (\citet{smith2017cyclical}) is employed throughout training. Along with this learning rate annealing strategy, we also use a Jaccard similarity coefficient between \(\mathbf{A}\) and \(\mathbf{A}^{*,o}\). First, \(\mathbf{A}\) is converted to a binary matrix \(\mathbf{A}^b\) such that any value \(\geq 0.5\) becomes \(1\) and \(< 0.5\) becomes \(0\). Then, the metric is calculated as:
\begin{equation}
    JSC_A = \frac{|\mathbf{A}^b \cap \mathbf{A^{*,o}}|}{|\mathbf{A}^b \cup \mathbf{A^{*,o}}|}
\end{equation}
While using cross-entropy as the metric for gradient descent helps in smooth loss reduction, it has certain limitations, (1) it is hard to know what a low cross-entropy loss is, and hence (2) knowing how long to train for is not straightforward. Having an upper-bound at \(1\), \(JSC_A\) helps bypass this problem. Training is done so that \(JSC_A\) reaches at least \(0.8\). As such, a training session takes \(\sim 21\) hours. All implementation is done in Python version 3.8.10 and a GPU cuda version 11.2.

\begin{table}
    \caption{Dictionary of all parameters used for the methodology and evaluations}
    \centering
    \renewcommand{\arraystretch}{1.2}
    \begin{tabular}{*{16}{c}} % 17 columns
    \hline
    \textbf{\(T\)} & \textbf{\(B\)} & \textbf{\(x_{lim}, y_{lim}\)} & \textbf{\(h\)} & \textbf{\(d_k\)} & \textbf{\(d_v\)} & \textbf{\(d_{ffn}\)} & \textbf{\(n_l\)} & \textbf{\(n_h\)} & \textbf{\(dt\)} & \textbf{\(\sigma_q\)} & \textbf{\(\sigma_r\)} & \textbf{\(\tau\)} & \textbf{\(a\)} \\
    \hline
    100 & 20 & 30 & 128 & 16 & 16 & 1024 & 6 & 8 & \(10^{-1}\) & \(10^{-3}\) & \(10^{-3}\) & 10 & 2 \\
    \hline
    \end{tabular}
    \label{tab:method_params}
\end{table}

\textbf{Data parameters and tasks:}
A patch of \(x_{lim} \times y_{lim}\) pixels is taken, and to its ground truth, corruptions are made to simulate measurements as described above. Different scenarios called tasks are defined containing the initializations of the parameters used for the said corruptions as described in Table \ref{tab:tasks}. Task \(\phi\) is a baseline task with no noise whatsoever added to it. Tasks A, B and C mildly introduce two of the three types of corruption. Tasks 1, 2, 3 and 4 are more realistic cases.

\begin{table}
    \caption{Task initialization for comparison}
    \centering
    \renewcommand{\arraystretch}{1.2}
    \begin{tabular}{c*{8}{c}}
        \hline
        & \(\phi\) & A & B & C & 1 & 2 & 3 & 4 \\
        \hline
        \(\sigma_m\) & 0 & 0 & 1 & 1 & 1 & 1 & 3 & 3 \\
        \(p_{fn}\) & 0 & \(10^{-1}\) & \(10^{-1}\) & 0 & \(10^{-1}\) & \(10^{-1}\) & \(10^{-1}\) & \(10^{-1}\) \\
        \(\lambda_{fp}\) & 0 & \(5\times 10^{-5}\) & 0 & \(5\times 10^{-5}\) & \(5\times 10^{-5}\) & \(2.5\times 10^{-4}\) & \(2.5\times 10^{-4}\) & \(3.3\times 10^{-4}\) \\
        \hline
    \end{tabular}
    \label{tab:tasks}
\end{table}

\section{Results}
\label{sec:results}
We first examine\footnote{All scores in Tables \ref{tab:tgospa_summary}, \ref{tab:mid_density_res} and \ref{tab:low_density_res} are reported as mean of 3 runs. TGOSPA scores are reported as mean \(\pm\) standard deviation of 3 runs. Other metrics showed negligible variability and are omitted for brevity.} the TGOSPA scores for ABHA and MHT across all tasks in medium and low density scenarios in Table \ref{tab:tgospa_summary}. TGOSPA captures location errors, missed and false detections, identity switches, and track fragmentation. ABHA consistently achieves lower TGOSPA scores than MHT across most tasks, particularly under medium density conditions. This indicates that ABHA makes more accurate overall trajectory associations. The only exception to this pattern is observed in idealized baseline tasks, where MHT performs better. This is likely due to the simplicity of the setting that aligns well with the greedy association strategy of MHT. This observation highlights that ABHA favors selective, high-confidence associations. This conservative strategy results in fewer false positives, which are heavily penalized by TGOSPA, thus resulting in a better performance.

\begin{table}[ht]
\caption{TGOSPA comparison between ABHA and MHT across tasks in medium and low density settings (\(\text{SNR} = 4\)).}
\centering
\renewcommand{\arraystretch}{1.2}
\begin{tabular}{ccccc}
\hline
\multirow{2}{*}{\textbf{Task}} & \multicolumn{2}{c}{\textbf{Medium Density}} & \multicolumn{2}{c}{\textbf{Low Density}} \\
                               & \textbf{ABHA} & \textbf{MHT}                & \textbf{ABHA} & \textbf{MHT} \\
\hline
\(\phi\) & \(1.294 \pm 0.114\) & \(0.837 \pm 0.006\) & \(2.312 \pm 0.012\) & \(1.283 \pm 0.003\) \\
\hline
A        & \(2.174 \pm 0.017\) & \(2.085 \pm 0.013\) & \(3.476 \pm 0.102\) & \(3.109 \pm 0.023\) \\
\hline
B        & \(2.222 \pm 0.103\) & \(2.296 \pm 0.003\) & \(3.199 \pm 0.130\) & \(1.632 \pm 0.074\) \\
\hline
C        & \(1.763 \pm 0.011\) & \(1.815 \pm 0.002\) & \(3.000 \pm 0.115\) & \(2.155 \pm 0.101\) \\
\hline
1        & \(2.111 \pm 0.006\) & \(2.390 \pm 0.034\) & \(2.105 \pm 0.037\) & \(2.884 \pm 0.006\) \\
\hline
2        & \(2.612 \pm 0.041\) & \(4.307 \pm 0.101\) & \(2.040 \pm 0.022\) & \(3.746 \pm 0.041\) \\
\hline
3        & \(2.313 \pm 0.088\) & \(4.528 \pm 0.015\) & \(2.136 \pm 0.103\) & \(5.154 \pm 0.127\) \\
\hline
4        & \(2.116 \pm 0.094\) & \(5.743 \pm 0.103\) & \(2.114 \pm 0.092\) & \(5.382 \pm 0.133\) \\
\hline
\end{tabular}
\label{tab:tgospa_summary}
\end{table}

For a more detailed understanding of the performance behavior, we compare the performance of both methods in terms of the other metrics defined in Section \ref{sec:perf_metrics} (Tables \ref{tab:mid_density_res} and \ref{tab:low_density_res}). ABHA generally scores higher in link precision \(P_L\) and comparable or slightly lower in link recall \(R_L\) relative to MHT. This indicates that while ABHA misses some true links, it is more accurate when it indeed decides to link. Similarly, its $F1_L$ and $JSC_L$ scores are competitive, indicating that ABHA tries to construct viable trajectories and not just avoid errors.

Interestingly, we also observe that ABHA consistently exhibits lower recall, both in terms of its points (\(R_P\)) and its links (\(R_L\)). This reflects its tendency to avoid uncertain associations. This trade-off could, again, be considered as a residual artifact of its conservative strategy, which proves advantageous in cluttered or noisy scenarios, where aggressive linking might lead to compounding errors.

Thus, we do not position ABHA as a universal outperformer, but as a conservative yet accurate tracker particularly suited for cluttered regimes. This helps us provide insight into how combining explicit filtering with a learned hypothesis-pruning strategy could offer better performance and adaptability in complex tracking scenarios.

\begin{table}[ht]
\caption{Comparison of ABHA and MHT on a patch of Viruses dataset for \(\text{SNR} = 4\) and medium density.}
\centering
\renewcommand{\arraystretch}{1.2}
\begin{tabular}{ccccccccc}
\hline
\textbf{Task} & \textbf{Method} & \(R_P\) & \(P_P\) & \(JSC_P\) & \textbf{\(P_L\)} & \textbf{\(R_L\)} & \textbf{\(F1_L\)} & \textbf{\(JSC_L\)} \\
\hline
\multirow{2}{*}{\(\phi\)} & ABHA & 0.974 & 0.974 & 0.948 & 0.957 & 0.908 & 0.932 & 0.466 \\
                          & MHT  & 0.990 & 0.992 & 0.996 & 0.977 & 0.994 & 0.986 & 0.493 \\
\hline
\multirow{2}{*}{A} & ABHA & 0.912 & 0.999 & 0.853 & 0.990 & 0.798 & 0.896 & 0.448\\
                   & MHT  & 0.943 & 0.984 & 0.845 & 0.947 & 0.838 & 0.889 & 0.445\\
\hline
\multirow{2}{*}{B} & ABHA & 0.907 & 0.994 & 0.823 & 0.772 & 0.643 & 0.832 & 0.411\\
                   & MHT  & 0.917 & 0.990 & 0.864 & 0.933 & 0.803 & 0.863 & 0.432\\
\hline
\multirow{2}{*}{C} & ABHA & 0.872 & 0.988 & 0.994 & 0.983 & 0.754 & 0.943 & 0.564\\
                   & MHT  & 0.998 & 0.942 & 0.883 & 0.939 & 0.994 & 0.966 & 0.483\\
\hline
\multirow{2}{*}{1} & ABHA & 0.914 & 0.988 & 0.876 & 0.943 & 0.677 & 0.877 & 0.534\\
                   & MHT  & 0.928 & 0.989 & 0.855 & 0.900 & 0.786 & 0.839 & 0.419\\
\hline
\multirow{2}{*}{2} & ABHA & 0.923 & 0.989 & 0.865 & 0.976 & 0.705 & 0.869 & 0.539\\
                   & MHT  & 0.990 & 0.873 & 0.804 & 0.815 & 0.867 & 0.840 & 0.420 \\
\hline
\multirow{2}{*}{3} & ABHA & 0.884 & 0.935 & 0.823 & 0.882 & 0.564 & 0.773 & 0.530\\
                   & MHT  & 0.979 & 0.823 & 0.720 & 0.612 & 0.647 & 0.629 & 0.315\\
\hline
\multirow{2}{*}{4} & ABHA & 0.845 & 0.902 & 0.811 & 0.834 & 0.561 & 0.784 & 0.511\\
                   & MHT  & 0.979 & 0.812 & 0.681 & 0.544 & 0.532 & 0.546 & 0.278\\
\hline
\end{tabular}
\label{tab:mid_density_res}
\end{table}

\begin{figure}
    \centering
    \includegraphics[width=\linewidth]{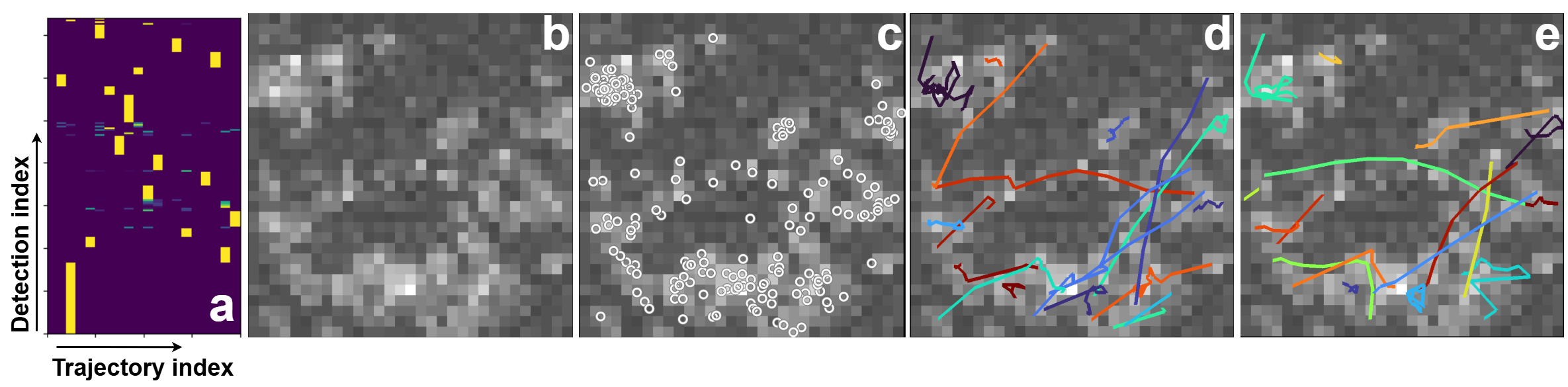}
    \caption{Visualization results of task 2 on medium density scenario. (a) predicted association matrix. (b) maximum intensity projection of time accumulated raw image. (c) time accumulated measurements. (d) time accoumulated ground-truth trajectories. (e) time accumulated predicted trajectories.}
    \label{fig:benchmark}
\end{figure}

\begin{table}[ht]
\caption{Comparison of ABHA and MHT on a patch of Viruses dataset for \((SNR = 4)\) and low density.}
\centering
\renewcommand{\arraystretch}{1.2}
\begin{tabular}{ccccccccc}
\hline
\textbf{Task} & \textbf{Method} & \(R_P\) & \(P_P\) & \(JSC_P\) & \textbf{\(P_L\)} & \textbf{\(R_L\)} & \textbf{\(F1_L\)} & \textbf{\(JSC_L\)}\\
\hline
\multirow{2}{*}{\(\phi\)} & ABHA & 0.902 & 0.944 & 0.820 & 0.811 & 0.748 & 0.778 & 0.389\\
                          & MHT  & 0.964 & 0.956 & 0.946 & 0.922 & 0.922 & 0.922 & 0.461\\
\hline
\multirow{2}{*}{A} & ABHA & 0.839 & 0.847 & 0.757 & 0.790 & 0.621 & 0.696 & 0.348\\
                     & MHT  & 0.911 & 0.785 & 0.645 & 0.823 & 0.767 & 0.794 & 0.397\\
\hline
\multirow{2}{*}{B} & ABHA & 0.813 & 0.930 & 0.690 & 0.770 & 0.553 & 0.644 & 0.322\\
                     & MHT  & 0.955 & 0.981 & 0.911 & 0.979 & 0.913 & 0.945 & 0.472 \\
\hline
\multirow{2}{*}{C} & ABHA & 0.875 & 0.875 & 0.766 & 0.774 & 0.699 & 0.735 & 0.367\\
                     & MHT  & 0.991 & 0.853 & 0.748 & 0.900 & 0.961 & 0.930 & 0.465\\
\hline
\multirow{2}{*}{1} & ABHA & 0.839 & 0.931 & 0.881 & 0.877 & 0.544 & 0.671 & 0.533\\
                     & MHT  & 0.973 & 0.858 & 0.772 & 0.848 & 0.815 & 0.831 & 0.416\\
\hline
\multirow{2}{*}{2} & ABHA & 0.843 & 0.675 & 0.499 & 0.769 & 0.549 & 0.641 & 0.540\\
                     & MHT  & 0.973 & 0.661 & 0.483 & 0.619 & 0.709 & 0.661 & 0.330\\
\hline
\multirow{2}{*}{3} & ABHA & 0.874 & 0.843 & 0.885 & 0.812 & 0.534 & 0.644 & 0.381\\
                     & MHT  & 0.973 & 0.568 & 0.397 & 0.488 & 0.602 & 0.539 & 0.270\\
\hline
\multirow{2}{*}{4} & ABHA & 0.887 & 0.932 & 0.843 & 0.932 & 0.441 & 0.599 & 0.392 \\
                     & MHT  & 0.964 & 0.532 & 0.358 & 0.357 & 0.437 & 0.393 & 0.196 \\
\hline
\end{tabular}
\label{tab:low_density_res}
\end{table}

\section{Conclusion}
\label{sec:conclusion}
We propose a hybrid multiple particle tracking framework that integrates the self-attention mechanism of the transformer architecture and Bayesian filtering. By casting data association as a label prediction problem, and hence using the transformer encoder to obtain association scores, we effectively prune the combinatorially large hypothesis space before applying Kalman filtering. This modularity of learning-based and model-based components allows our method to maintain robustness in noisy and cluttered environments while maintaining interpretability, say, in the form of the ability to evaluate the likelihood of a trajectory through the predicted association matrix. As such, we demonstrate that the proposed approach achieves improved tracking accuracy according to several metrics, particularly in high-noise settings where traditional methods degrade. These results support the idea that attention contributes effectively at capturing relational structure of the data, making it well suited for complex association tasks that benefit from learned priors.

\section{Limitations and Future Scope}
\label{sec:lims}
Our training and evaluation are limited to a single
type of particle motion, a mix of Brownian and directed dynamics typical of virus trafficking. While this dynamic is of high biological relevance, it would be insightful to test our method on a broader set of motion types and mixtures. Moreover, due to computational constraints, we restrict training and evaluation to (30\(\times\)30) patches of the full field of view, allowing the tracking of \(\sim\) 20 particles at a time. This prevents the study from acting as a full benchmark but a comparison within a limited spatial and motion context. Nevertheless, it enables the precise study of the benefit of our hybrid approach when compared to conventional Bayesian techniques. Furthermore, MHT consistently has better \(R_P\) and \(R_L\) scores. We also see the difference in performance between both the methods compared decrease on moving from a medium-density to a low-density scenario. Analyzing and understanding the reasoning for it is an interesting and necessary extension strategy to update the data structure and framework design. This is required to have a method that could potentially outperform MHT in all the performance metrics discussed. The primary goal of the future scope of this study is to address these limitations. We will move towards a more generalizable training of this tracking strategy to infer on different types of biologically relevant movements. In addition, strategies will be implemented to work in a larger field of view for a more standardized comparison.

\begin{ack}
    The project leading to this publication has received funding from France 2030, the French Government program managed by the French National Research Agency (ANR-16-CONV-0001) and from Excellence Initiative of Aix-Marseille University - A*MIDEX.
\end{ack}

\bibliography{ref}

\newpage
\appendix

\section{Prerequisites}
\label{sec:prerequisites}
\subsection{Self-attention learns the relationship between each element of the input set}
In this section we describe the theory behind self-attention, introduced by \citet{vaswani2017attention}. Consider \(\mathbf{E} \in \mathbb{R}^{n \times h}\) to be the input data structure for some hidden dimension \(h\). \(\mathbf{Q}, \mathbf{K} \in \mathbb{R}^{n \times d_k}\) and \(\mathbf{V} \in \mathbb{R}^{n \times d_v}\) are projections of \(\mathbf{E}\) in different learned subspaces such that:
\begin{equation}
\begin{cases}
\mathbf{Q} = \mathbf{E}.\mathbf{W}_q\\
\mathbf{K} = \mathbf{E}.\mathbf{W}_k\\
\mathbf{V} = \mathbf{E}.\mathbf{W}_v
\end{cases}
\end{equation}
where \(\mathbf{W}_q, \mathbf{W}_k \in \mathbb{R}^{h \times d_k}\) and  \(\mathbf{W}_v \in \mathbb{R}^{h \times d_v}\) are learned weights. Two of these projections \(\mathbf{Q}\) and \(\mathbf{K}\) are used to calculate a score between each element of \(\mathbf{E}\) and yet another projection \(\mathbf{V}\) is used to carry the information from the input that would get mixed based on the aforementioned score. This is done by taking the softmax of the scaled-dot-product between \(\mathbf{Q}\) and \(\mathbf{K}^T\) (\(\in \mathbb{R}^{n \times n}\)) and multiplying it with \(\mathbf{V}\), such as:
\begin{equation}
    \text{attention}(\mathbf{Q}, \mathbf{K}, \mathbf{V}) = \text{softmax}(\frac{\mathbf{Q}.\mathbf{K}^T}{\sqrt{d_k}}).\mathbf{V} \in \mathbb{R}^{n \times d_v}
\end{equation}
In multi-head attention, we concatenate multiple attention heads before applying a final projection. Consider \(n_h = \frac{h}{d_v} \in \mathbb{N}\) attention heads (which requires \(h\) to be divisible by \(d_v\), and hence \(n_h\)), and the input projections of the \(j^{th}\) attention-head to be \(\mathbf{Q}_j\), \(\mathbf{K}_j\) and \(\mathbf{V}_j\). Hence the output of this multi-head attentions is:
\begin{equation}
    MHA = \text{concat}(\text{attention}(\mathbf{Q}_j, \mathbf{K}_j, \mathbf{V}_j), \forall j \in [1, n_h]) \in \mathbb{R}^{n \times h}
\end{equation}

\subsection{Transformer encoder is a neural network that uses self-attention}
Suppose that the transformations undergone by the input \(\mathbf{E}\) in the multi-head self-attention layer lead to an output \(MHA(\mathbf{E}) \in \mathbb{R}^{n \times h}\). Subsequently, the following transformations take place in an encoder layer,
\begin{equation}
    \mathbf{RO} = \mathbf{E} + MHA \in \mathbb{R}^{n \times h}
\end{equation}
\begin{equation}
    \mathbf{NO} = \text{LayerNorm}(\mathbf{RO}) \in \mathbb{R}^{n \times h}
    \label{eq:ln}
\end{equation}

\begin{equation}
    \mathbf{FO} = \phi(\mathbf{NO}.\mathbf{W}_1 + \mathbf{b}_1)\mathbf{W}_2 + \mathbf{b}_2 \in \mathbb{R}^{n \times h}
\end{equation}
where a non-linearity activation function \(\phi\) is applied between the layers; \(\mathbf{W}_1 \in \mathbb{R}^{h \times d_{ffn}}\),  \(\mathbf{W}_2 \in \mathbb{R}^{d_{ffn} \times h}\), \(\mathbf{b}_1 \in \mathbb{R}^{n \times d_{ffn}}\) and \(\mathbf{b}_2 \in \mathbb{R}^{n \times h}\). In eq. \ref{eq:ln}, layer normalization (\citet{ba2016layer}) helps stabilize training.
\begin{equation}
    \mathbf{RO}_{ffn} = \mathbf{NO} + \mathbf{FO} \in \mathbb{R}^{n \times h}
\end{equation}
\begin{equation}
    \text{Final output} = \text{LayerNorm}(\mathbf{RO}_{ffn}) \in \mathbb{R}^{n \times h}
\end{equation}
If we combine all the transformations in an encoder layer to call it \(\mathscr{E}\), the output of one encoder layer is \(\mathscr{E}(\mathbf{E})\). In an encoder, there are multiple such encoder layers stacked one after the other. So, for an encoder with \(n_l\) encoder layers, the output can be called \(\mathscr{E}_{n_l}(\mathbf{E}) \in \mathbb{R}^{n \times h}\).

\section{Derivation of our process noise covariance matrix in Kalman filtering}
\label{sec:process_matrix}
For continuous time in one dimension, we can describe our state as a function of time as:

\begin{equation}
    \mathbf{x}(t) = \begin{bmatrix}
			x(t)\\
			v(t)
\end{bmatrix}
\end{equation}

The state transition equation can be written as:
\begin{equation}
    \frac{d}{dt}\begin{bmatrix}x(t)\\v(t)\end{bmatrix} = \begin{bmatrix}0 & 1\\0 & 0\end{bmatrix}\begin{bmatrix}x(t)\\v(t)\end{bmatrix} + \begin{bmatrix}0\\1\end{bmatrix}a(t)
\end{equation}

where $a(t)$ is the process noise that arises from unknown random acceleration. Let $a(t)$ be a zero-mean white noise process so $a(t)$ and $a(t')$ for different time-points $t$ and $t'$ are uncorrelated. Hence, noise covariance can be modeled using the Dirac-delta function:

\begin{equation}
    \mathbb{E}[a(t)a(t')] = \sigma_q^2\delta(t-t')
\end{equation}
where 
\begin{equation}
    \delta(t-t') = \begin{cases}\infty, & t=t'\\ 0, & \text{else}\end{cases}
\end{equation}
So, when $t \neq t'$, noise autocorrelation is zero, meaning process noise at two different time steps is not correlated. This affects velocity and position differently

\begin{equation}
    v(t) = v(0) + \int_0^ta(\tau)d\tau
\end{equation}
\begin{equation}
    x(t) = x(0) + v(0)t + \int_0^t\int_0^sa(\tau)d\tau ds
\end{equation}

As such, the velocity-velocity covariance is determined as,
\begin{equation}
    \mathbb{E}[v(t)v(t')] = \int_0^t \int_0^{t'} \mathbb{E} [a(\tau)a(\tau ')]d\tau 'd\tau
\end{equation}

\begin{equation}
     \mathbb{E}[v(t)v(t')] = \sigma_q ^2 \int_0^t \int_0^{t'} \delta(\tau - \tau ')d\tau 'd\tau
\end{equation}
Now,
\begin{equation}
    \int_0^{t'} \delta(\tau - \tau ')d\tau ' = \begin{cases}1, & \tau \in [0,t']\\ 0, & \text{else}\end{cases} = h(t' - \tau) \text{ (let)}
\end{equation}
So,
\begin{equation}
    \mathbb{E}[v(t)v(t')] = \sigma_q^2\int_0^th(t'-\tau)d\tau
\end{equation}

\begin{equation}
   \mathbb{E}[v(t)v(t')] = \sigma_q^2min(t, t')
\end{equation}

at $t=t'$,
\begin{equation}
    \mathbb{E}[v^2(t)] = \sigma_q^2 t
\end{equation}

Hence, the velocity-velocity covariance is affected such that a linear increase in time increases the covariance linearly. Similarly, position-velocity covariance has a quadratic dependence on time,
\begin{equation}
    \mathbb{E}[x(t)v(t)] = \frac{\sigma_q^2t^2}{2}
\end{equation}
and position-position covariance has a cubic dependence on time.
\begin{equation}
    \mathbb{E}[x^2(t)] = \frac{\sigma_q^2t^3}{3}
\end{equation}

Consider, for the state $\mathbf{x}(t)$, a process noise covariance matrix denoted by $\mathbf{Q}(t) = \begin{bmatrix}q_{xx}(t) & q_{xv}(t)\\ q_{vx}(t) & q_{vv}(t)\end{bmatrix}$, where $q_{xx}$ is the uncertainty of position with respect to itself, $q_{vv}$ is the uncertainty of velocity with respect to itself, and $q_{xv}$ and $q_{vx}$ are uncertainties of position with respect to velocity and vice versa.
\begin{equation}
    \mathbf{Q}(t) = \sigma_q^2\begin{bmatrix}\frac{t^3}{3} & \frac{t^2}{2}\\ \frac{t^2}{2} & t\end{bmatrix}
\end{equation}

Hence, for a discrete time sampling interval $dt$, the process noise matrix in 2-dimensions becomes,
\begin{equation}
    \mathbf{Q} = \sigma_q^2\begin{bmatrix}\frac{dt^3}{3} & \frac{dt^2}{2} & 0 & 0\\ \frac{dt^2}{2} & dt & 0 & 0\\ 0 & 0 & \frac{dt^3}{3} & \frac{dt^2}{2}\\ 0 & 0 & \frac{dt^2}{2} & dt\end{bmatrix}
\end{equation}

Thus, we obtain the process noise covariance matrix used in our model.

\section{Generalized Optimal Sub-Pattern Assignment}
\label{sec:gospa}

The GOSPA metric evaluates multi-object tracking accuracy at a single time step, incorporating both localization and cardinality errors. The GOSPA distance of order \( p \geq 1 \) with cutoff \( c > 0 \) is defined as,
\begin{equation}
    d_p^c(\mathbf{X}_t, \hat{\mathbf{X}}_t) = \left( \min_{\pi \in \Pi} \sum_{i=1}^{k} \min\left( \|\mathbf{x}_t^i - \chi_t^{\pi(i)}\|^p, c^p \right) + \frac{c^p}{2}(|\mathbf{X}_t| + |\hat{\mathbf{X}}_t| - 2k) \right)^{1/p}
\end{equation}
where, \( \Pi \) is the set of all injective assignments between the smaller of the two sets. Moreover, \(\pi(i) = j\) means that the \(i^{th}\) index of \(\mathbf{X}\) has matched with the \(j^{th}\) index of \(\hat{\mathbf{X}}\). Further, \( k = \min(|\mathbf{X}|, |\hat{\mathbf{X}}|) \). The first term measures the localization error between the matched points, while the second penalizes the unmatched ground-truth or predicted detections.

\begin{table}
    \caption{Dictionary of parameters used for performance metrics}
    \centering
    \begin{tabular}{cc}
    \hline
      \(d_{\phi}\)   & 3  \\
       \(c\)  & \(5\)\\
       \(p\) & \(2\)\\
       \hline
    \end{tabular}
    \label{tab:perf_params}
\end{table}

\end{document}